\documentclass[conference]{IEEEtran}
\IEEEoverridecommandlockouts
\usepackage{cite}
\usepackage[bookmarks=false]{hyperref}
\usepackage{amsmath,amssymb,amsfonts}
\usepackage{algorithmic}
\usepackage{graphicx}
\usepackage[caption=false]{subfig}
\usepackage{textcomp}
\usepackage{siunitx}
\usepackage{multirow}
\usepackage{hyperref}
\usepackage{gensymb}
\usepackage{booktabs}
\usepackage[table]{xcolor}
\def\BibTeX{{\rm B\kern-.05em{\sc i\kern-.025em b}\kern-.08em
    T\kern-.1667em\lower.7ex\hbox{E}\kern-.125emX}}

\begin{document}

\title{Stone Soup Multi-Target Tracking Feature Extraction For Autonomous Search And Track In Deep Reinforcement Learning Environment
\thanks{Jan-Hendrik Ewers is a PhD candidate on secondment at Leonardo Electronics funded by EPSRC grant number \texttt{EP/T517896/1-312561-05}. Joe Gibbs is a sponsored PhD candidate with Leonardo Electronics.}
}
\author{\IEEEauthorblockN{Jan-Hendrik Ewers$^{1}$}
\IEEEauthorblockA{\textit{Autonomous Systems and Connectivity} \\
\textit{University of Glasgow}\\
Glasgow, UK \\
j.ewers.1@research.gla.ac.uk}
\and
\IEEEauthorblockN{Joe Gibbs$^{1}$}
\IEEEauthorblockA{\textit{Autonomous Systems and Connectivity} \\
\textit{University of Glasgow}\\
Glasgow, UK \\
j.gibbs.1@research.gla.ac.uk}

\and
\IEEEauthorblockN{David Anderson$^{2}$}
\IEEEauthorblockA{\textit{Autonomous Systems and Connectivity} \\
\textit{University of Glasgow}\\
Glasgow, UK \\
dave.anderson@glasgow.ac.uk}
}

\maketitle

\begin{abstract}
Management of sensing resources is a non-trivial problem for future military air assets with future systems deploying heterogeneous sensors to generate information of the battlespace. Machine learning techniques including deep reinforcement learning (DRL) have been identified as promising approaches, but require high-fidelity training environments and feature extractors to generate information for the agent. This paper presents a deep reinforcement learning training approach, utilising the Stone Soup tracking framework as a feature extractor to train an agent for a sensor management task. A general framework for embedding Stone Soup tracker components within a Gymnasium environment is presented, enabling fast and configurable tracker deployments for RL training using Stable Baselines3. The approach is demonstrated in a sensor management task where an agent is trained to search and track a region of airspace utilising track lists generated from Stone Soup trackers. A sample implementation using three neural network architectures in a search-and-track scenario demonstrates the approach and shows that RL agents can outperform simple sensor search and track policies when trained within the Gymnasium and Stone Soup environment.
\end{abstract}


\section{Introduction}
Future airborne systems operating as a node in a \textit{system-of-systems} architecture will be required to utilise information gathered from heterogeneous sensors, capturing traditional observations of the environment using Radar and EO/IR sensors but also signals of opportunity. With an increase in the variety of sensing capabilities and advanced threats likely to be encountered, military platforms must effectively manage these tracking resources to maximise the information-gathering capability with constraints on signal processing and with some level of autonomy. The decision-making agent can be designed using traditional methods such as behaviour trees, partially-observable Markov decision processes (POMDP) or as a neural network trained through supervised or reinforcement learning.

The Stone Soup Python library \cite{thomas2017open,thomas2019stone} has been developed to provide an open-source, document-driven framework for developing novel state estimation algorithms for use in single and multi-target tracking along with sensor fusion and sensor management problems. Since it's initial release, the library has grown rapidly, incorporating new nonlinear estimation algorithms such as the adaptive kernel Kalman filter (AKKF) \cite{wright2023implementation,wright2024implementation} and expanding the supported scenarios to include orbital estimation, drone tracking and sensor management \cite{barr2022stone}. In it's current form, Stone Soup can be used to perform single or Monte Carlo simulations in state estimation problems. However, the component-based architecture provides an approachable implementation that could be utilised within wider simulation environments as a general tracking algorithm or, using the vernacular of machine learning, as a feature extractor. Feature extractors are methods that pre-process raw environment information using prior knowledge into a more useful representation for the agent. This task can be performed by algorithms such as state estimators for providing observable but unmeasured states, or using neural networks such as recurrent or convolutional networks to extract features from time series or image data.

Reinforcement learning is a machine learning technique akin to the learning approach of humans where an agent learns through interaction with an environment. The agent or policy takes a set of observations of the environment $\boldsymbol{s}_{k+1}$ and previous actions $\boldsymbol{a}_{k}$ and uses these to generate the next action $\boldsymbol{a}_{k+1}$. The agent is trained by selecting actions that maximise a user-defined reward function which incorporates positive reinforcement for desired behaviours and potentially negative penalties for undesired performance. RL is a popular approach to problems that feature multiple objectives, constraints, and high-dimensional observation or action spaces. The theory is that a suitably constructed agent is able to learn underlying structures in the observation data and relate this to a set of actions that will maximise the desired reward function. RL has been shown to learn novel approaches and outperform state-of-the-art algorithms in applications including wilderness search and rescue \cite{ewers2025deep}, drone racing \cite{Kaufmann2023} and abstract strategy games including chess and Go \cite{silver2018general}. A key result from \cite{Kaufmann2023} is that even simple network architectures, such as a MLP with a hidden component of $2\times{}128$ neurons, can excel if environment information is provided in a useful representation. While providing raw, unfiltered information should theoretically allow for the agent to extract the key information, in this case information about the area of interest including target states formatted into a track list, feature extractors are used to pre-process the observations to abstract complexity which could include frame or coordinate transformations, from the training process. For a Radar sensor management problem where an agent is required to scan cells and maximise the information of the scene, knowledge of current target states defined in the track list, along with an overall map of cells scanned will provide time-series information for the agent.  

Stone Soup has previously been used in RL problems for sensor management and drone tracking using tracker components and the Stable Baselines RL library \cite{de2023resources}. The former implementation is demonstrated in the Stone Soup documentation highlighting trackers utilised alongside the \textit{Tensorflow-Agents} library, though results are shown to be on par with traditional approaches. Recent work in \cite{Xiong2023} implemented a multi-agent reinforcement learning scenario using a basic tracking algorithm to generate observation state data. This work removed the association task by assuming wide separation of targets and assumed knowledge of the constant number of targets. These are limitations that can be easily be addressed by utilising components within the Stone Soup framework shown in Figure \ref{fig:MTT_Arch}. In this paper, an architecture for embedding Stone Soup tracker components as feature extractors within a Stable Baselines3 reinforcement learning framework using a Gymnasium environment is presented and demonstrated in an AESA Radar sensor management problem. The approach for implementing a MTT using Stone Soup components is discussed and additional interfaces required to leverage the Stable Baselines3 library for policy training are defined.
\subsection{Motivations for use of Stone Soup}
Stone Soup was envisaged to support researchers and engineers interested in developing novel state estimation algorithms, validating results using a benchmark of prior algorithms and for use in wider systems where knowledge of the internal workings of MTT algorithms would not be required \cite{barr2022stone}. The latter user base offers a broad scope for future implementations where tracking algorithms are critical for development of control systems and training of agents through reinforcement learning. The Stone Soup documentation currently provides an example implementation within a Tensorflow environment to train an agent in a sensor management problem, and an extension of this methodology to the widely-used and open-source Stable Baselines3 library \cite{raffin2021stable} and Gymnasium \cite{towers2024gymnasium} environments which it utilises would open up opportunities to researchers looking to utilise MTT algorithms within their training environments.

Current open-source libraries listed in Table \ref{tab:openSourceLibraries} including FilterPy, OpenKF and TrackerComponentLibrary \cite{crouse2017tracker} provide implementations for state estimation and tracking algorithms but are more suited to informed researchers or engineers with knowledge of the desired process and measurement models rather than those looking to include trackers into other fields. The vast array of tracker components also provides flexibility in the level of tracker implemented, with simpler and faster to run algorithms offering improved training times. The ability to change MTT objects within a Gymnasium environment also enables future \textit{curriculum learning} \cite{bengio2009curriculum} approaches where agents are trained in environments of increasing complexity to improve training times. TrackerComponentLibrary is the closest in feature set but is constructed with MATLAB functions rather than an object-oriented architecture with abstraction and inheritance. As such, more expertise is required to correctly implement larger and more complex tracking algorithms, and the lack of documentation further highlights the benefits of using Stone Soup.
\begin{table}[htbp]
\caption{Open-source state estimation and tracking libraries.}
\begin{center}
\begin{tabular}{|c|c|c|c|}
\hline
\textbf{Library}&\textbf{Language}&Link&\textbf{Reference}\\
\hline
Stone Soup&Python&\href{https://github.com/dstl/Stone-Soup}{Link}&\cite{thomas2017open,thomas2019stone,barr2022stone}\\
\hline
TrackerComponentLibrary&MATLAB&\href{https://github.com/USNavalResearchLaboratory/TrackerComponentLibrary}{Link}&\cite{crouse2017tracker}\\
\hline
FilterPy&Python&\href{https://github.com/rlabbe/filterpy}{Link}&\cite{labbe2018filterpy}\\
\hline
kalman&C++&\href{https://github.com/mherb/kalman}{Link}&\cite{labbe2018filterpy}\\
\hline
OpenKF&C++&\href{https://github.com/Al-khwarizmi-780/OpenKF}{Link}&\\
\hline
\end{tabular}
\label{tab:openSourceLibraries}
\end{center}
\end{table}
Due to Python being the dominant programming language used for reinforcement learning and ML or DL in general, a Python-based library would be the best choice for integration with existing RL libraries. Execution time is also a primary concern, as training reinforcement learning agents is to some extent a waiting game. FilterPy was found to be slow when integrated into a wider MTT algorithm. Stone Soup is the only library to provide full support for simulating platforms, sensors and implementation of component-based tracking algorithms with metric components built in. This makes it the ideal candidate for utilisation within Gymnasium reinforcement learning environments. The object-oriented structure also means that users can select the level of abstraction for the problem, using simpler, lower fidelity components that run faster or, if necessary, to re-implement slower functions to improve training times. Combined with this, an active user base and development team from the UK's Defence Science and Technology Laboratory (DSTL) and academia incentivise use within machine learning problems.

\section{Methodology}\label{sect:method}
\begin{figure}
    \centering
    \includegraphics[width=\linewidth]{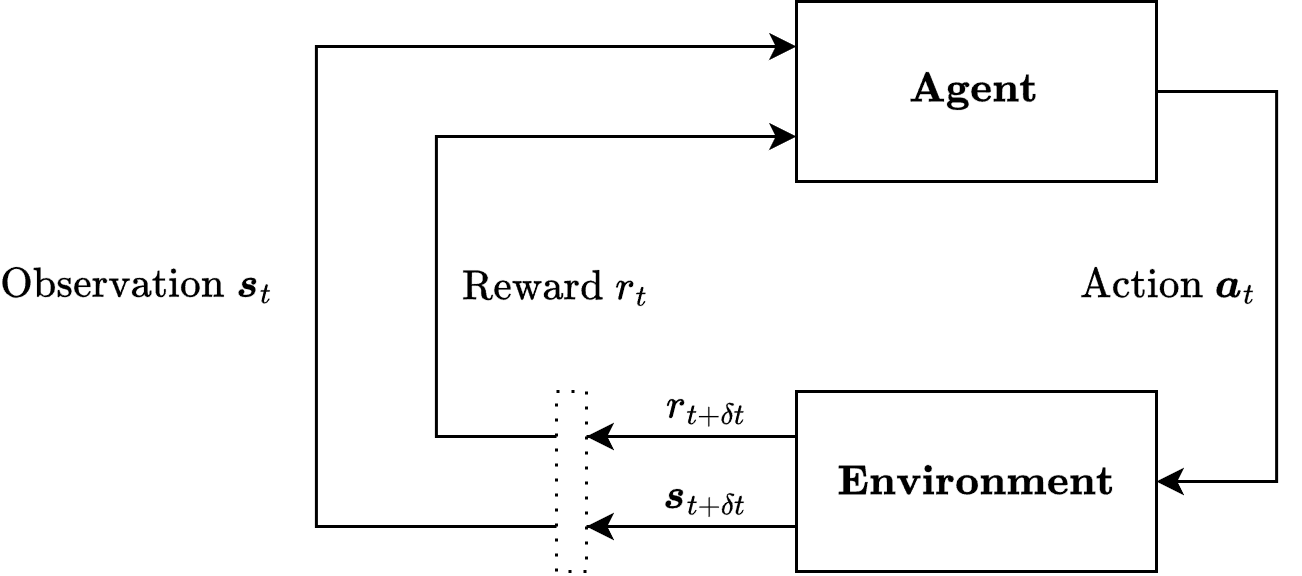}
    \caption{Reinforcement learning training process adapted from \cite{sutton1998reinforcement}. The agent interacts with the environment through a defined action set $\boldsymbol{a}_{t}$ based on the observation $\boldsymbol{s}_{t}$ and the reward $r_t$.}
    \label{fig:rl_training}
\end{figure}

Reinforcement learning requires two primary components for implementation. An agent, comprised of a suitably constructed neural network architecture with an input state featuring observable information either measured from the environment or processed with a feature extractor, and an environment through which the agent can interact with and which will generate the scenario data. Figure \ref{fig:rl_training} shows the general RL training process where the agent interacts with the environment through it's action, while receiving an observation state from the environment.
Gymansium \cite{towers2024gymnasium} is a fork of OpenAI Gym which provides a standard template for a RL environment which can be modified using wrappers for implementation in a wide range of problems. Fitting Stone Soup within a Gymnasium environment provides a base workflow for incorporating tracking components within an RL environment.
\begin{figure}
    \centering
    \includegraphics[width=\linewidth]{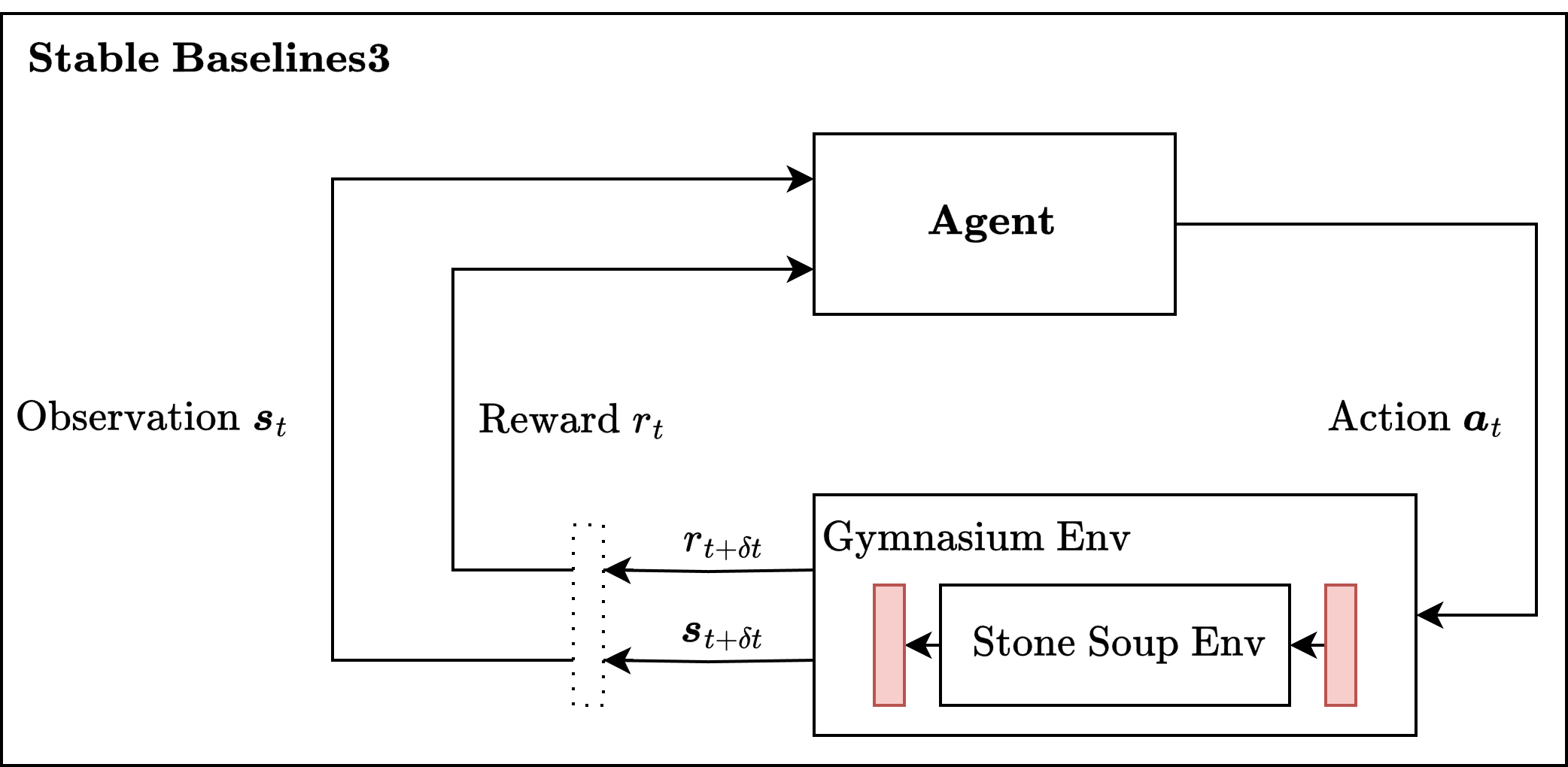}
    \caption{RL problem structure utilising Gymnasium environment containing Stone Soup components with suitable wrapper functions to integrate agent actions $\boldsymbol{a}_{k}$ with the sensor.}
    \label{fig:EnvArch}
\end{figure}
\subsection{Stone Soup Component Framework}
\begin{figure}
    \centering
    \includegraphics[width=\linewidth]{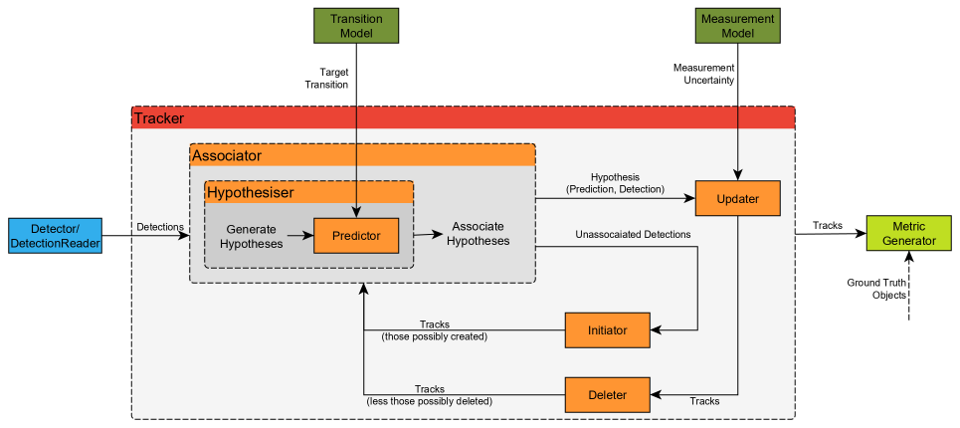}
    \caption{Stone Soup MTT component framework as provided in \cite{barr2022stone}.}
    \label{fig:MTT_Arch}
\end{figure}

The Stone Soup framework is composed of Algorithm and Enabling Components. Algorithm components are used to create the state estimator algorithm, composed of subclasses for each of the required subcomponents. Enabling components are defined to create objects within the environment and metrics such as GOSPA \cite{Rahmathullah2017} and Posaterior Cramer Rao Lower Bound (PCRLB) utilised truth data to assess performance of the state estimation algorithm. Objects can include platforms, sensors, detectors and sensor managers that are used to create a tracking scenario. The architecture of the algorithm components in the Stone Soup framework used to build a general multi-target tracker are presented in Figure \ref{fig:MTT_Arch}.
Table \ref{tab:stoneSoupComponents} details the Stone Soup algorithm components from Figure \ref{fig:MTT_Arch} used in the reinforcement learning environment.
\begin{table}[htbp]
\caption{Stone Soup Components used to implement MTT Feature Extractor}
\begin{center}
\begin{tabular}{|c|c|}
\hline
\textbf{Stone Soup Algorithm Component}&\textbf{Component Class}\\
\hline
Data Association&\textit{NearestNeighbour}\\
\hline
Initiator&\textit{SimpleMeasurementInitiator}\\
\hline
Predictor&\textit{KalmanPredictor}\\
\hline
Updater&\textit{UnscentedKalmanUpdater}\\
\hline
Hypothesiser&\textit{DistanceHypothesiser}\\
\hline
Deleter&\textit{CovarianceBasedDeleter}\\
\hline
Gater&\textit{DistanceGater}\\
\hline
\end{tabular}
\label{tab:stoneSoupComponents}
\end{center}
\end{table}
The \textit{DistanceHypothesiser} component utilises a suitable distance measure such as Kullback-Leibler divergence, Gaussian Hellinger or in this case, Mahalanobis distance. The supporting functions, referred to as \textit{Enabling Components} within the framework are presented in Table \ref{tab:stoneSoupComponents2}. Enabling components provide models for the sensors, platforms, ground truth simulation along with metrics for analysis of tracking performance. These metrics are not only useful for analysing the performance of reinforcement learning agents, but can be used as training epoch stop criteria if the covariance norm or GOPSA distance exceeds defined thresholds. 
\begin{table}[htbp]
\caption{Stone Soup supporting Metrics and Measures to implement MTT Feature Extractor}
\begin{center}
\begin{tabular}{|c|c|}
\hline
\textbf{Stone Soup Enabling Component}&\textbf{Component Function}\\
\hline
Measure&\textit{Mahalanobis}\\
\hline
\multirow{3}{4em}{  Metrics}&\textit{SumofCovarianceNormsMetric}\\&\textit{GOSPA}\\&\textit{SIAPMetrics}\\
\hline
Sensor&\textit{RadarRotatingBearingRange}$^{\ast}$\\
\hline
Platforms&\textit{FixedPlatform}$^{\ast}$\\
\hline
Simulators&\textit{MultiTargetGroundTruthSimulator}\\
\hline
\multicolumn{2}{l}{$^{\mathrm{\ast}}$Customisation required for implementation.}
\end{tabular}
\label{tab:stoneSoupComponents2}
\end{center}
\end{table}
Stone Soup provides a repository of target generators for simulations that can be randomised to enable environment resets at each training epoch. Since the observations generated from the Stone Soup \textit{Sensor} component rely upon the agent outputs, the renormalised actions $\boldsymbol{a}_{k}$ representing the desired bearing and elevation pointing commands for the Radar. Stone Soup \textit{Sensor} components have an action component which can update the dwell angle of the sensor to perform this. External actions can therefore be used to update the sensor configuration prior to generating the measurements at the current time step.
\begin{figure*}
    \centering
    \includegraphics[width=0.8\linewidth]{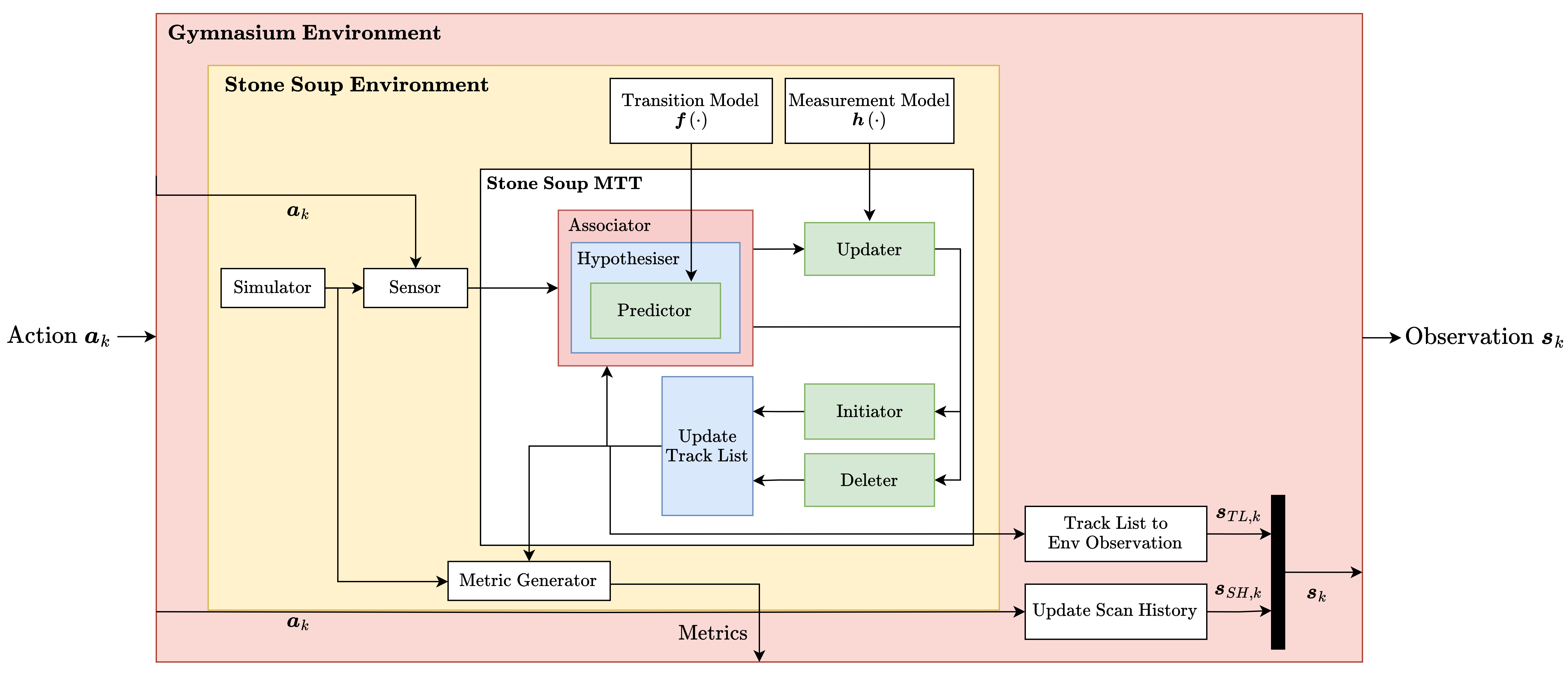}
    \caption{Gymnasium \cite{towers2024gymnasium} environment construction utilising Stone Soup components for ground truth generation, multi-target tracker implementation and metrics generation.}
    \label{fig:env_arch}
\end{figure*}
The Stone Soup environment generates the updated track list and state covariances at each time step. To generate the full observation state $\boldsymbol{s}_{k}$ at the current time step, the track list information must be converted to the observation state form $\boldsymbol{s}_{TL_k}$ and concatenated with the updated scan or look-history $\boldsymbol{s}_{SH_k}$ to form $\boldsymbol{s}_{k}$. These functions are built within the Gymnasium wrapper to enable implementation within the Stable Baselines3 framework.
\subsection{Wrapper Functions for Stone Soup implementation}
Two main wrapper functions are required to interface a Stone Soup MTT with the Gymnasium environment utilised for training using Stable Baselines3. The first is the construction of the observation state $\boldsymbol{s}_{k}$ from information generated by the Stone Soup components. This may or may not be limited to just the track information, but must be concatenated to provide all information in the correct form for the agent to interpret. Additionally, any metrics required by the reward function must be generated and provided to the Gymnasium environment. The second is any transformation required to map the agent action to the command to the sensor object. For the 3D search-and-track problem, the required transformation is shown in (\ref{eq:act_1}) and (\ref{eq:actionToBearings}).
\subsection{Current Limitations}
At present, only a discrete action space for a active sensor in the sensor management problem is suitable for use with the Stone Soup framework. Specifically for the \textit{RadarRotatingBearingRange} sensor component, providing an angular position command can cause a large number of possible actions to be generated which will all be simulated. As such, either a customised version of the sensor model, as performed in this study is required, or ground truth simulation and sensor modelling should be performed inside the Gymnasium environment but outside of the Stone Soup implementation.
\section{Demonstrative Sensor Management Problem}
The agent is trained in a representative Gymnasium environment simulating a typical engagement with a static observer platform that is tasked with searched the area of interest and tracking any targets identified. The AESA Radar generates measurements in a $9\degree$ field-of-view (FoV) with dwell centre defined by the agent. Agent actions are considered to be instantaneous with no penalty for rotating the sightline away from the current line-of-sight. For demonstration of a basic MT algorithm, a fixed set of targets are birthed at the start of the simulation $t=t_{0}$ with deaths at $t=t_{f}$.
\begin{table}[htbp]
\caption{MTT Parameters utilising Stone Soup components in Table \ref{tab:stoneSoupComponents}}
\begin{center}
\begin{tabular}{|c|c|}
\hline
\textbf{Parameter}&\textbf{Value}\\
\hline
$T_{s}$&$0.005\textrm{s}$\\
\hline
Deleter Threshold&$5000$\\
\hline
\end{tabular}
\label{tab:MTTparams}
\end{center}
\end{table}
\subsection{Multiple Target Tracker Feature Extraction}
The RL agent has access to measurement information from the Radar sensor which contains information on the current tracks in the area-of-interest. However, there are severe limitations with providing raw data directly to the agent. Neural networks train best with normalised inputs and outputs, with values in the interval $[0,1]$ reducing the sensitivity of the network to changes in input. This process can be easily performed on Radar returns and would form the transformation directly prior to entry to the agent. Radar observations are also noise-corrupted and un-correlated. Measurements won't provide context for the number of targets since the agent will see a list of possible observations without any notion of the number of separate targets. Realistic measurements will also be generated from scene clutter which will provide misinformation to the agent and result in increased training time and poorer performance. Critically, the limited beamwidth of the Radar and reliance on agent actions will mean that only measurements within the current field of view will be generated, providing no information about the remainder of the area of interest. A MTT algorithm will propagate tracks without measurements and maintain information for the agent in between sector revisits. Raw Radar returns will provide information only when a sector is scanned, and without maintained track information the agent will not be able to utilise information of targets outside the current field-of-view. Instead, the MTT algorithm will maintain a track list of estimated targets, kinematic states and covariances which will assist the agent in selecting which actions will maximise the information available, or the reduction in overall covariance. A constant-velocity target model \cite{li2003survey}, implemented using the \textit{LinearGaussianTransitionModel} component is used as the transition model with measurements generated using a custom implementation of the \textit{RadarRotatingBearingRange} to enable a moving sightline. Since this work was undertaken, a 3D variant \textit{RadarRotatingBearingElevationRange} has been released which may mitigate some of the required customisations.
\subsection{Definition of the agent and environment interfaces}
The agent is required to select bearing and elevation angle commands to search the area of interest for new targets and update existing tracks to maximise the scenario information. To complete this effectively, the agent requires information of the current track list generated by the Stone Soup MTT object, and the history of previous scans. These obsevation states are denoted as $\boldsymbol{s}_{TL}$ and $\boldsymbol{s}_{SH}$ respectively.
Modelling assumptions used for the scan history and further details of the simulation can be found in \cite{ewers2025multi}.
\subsection{Action Space}
\begin{figure}
    \centering
    \includegraphics[width=0.8\linewidth]{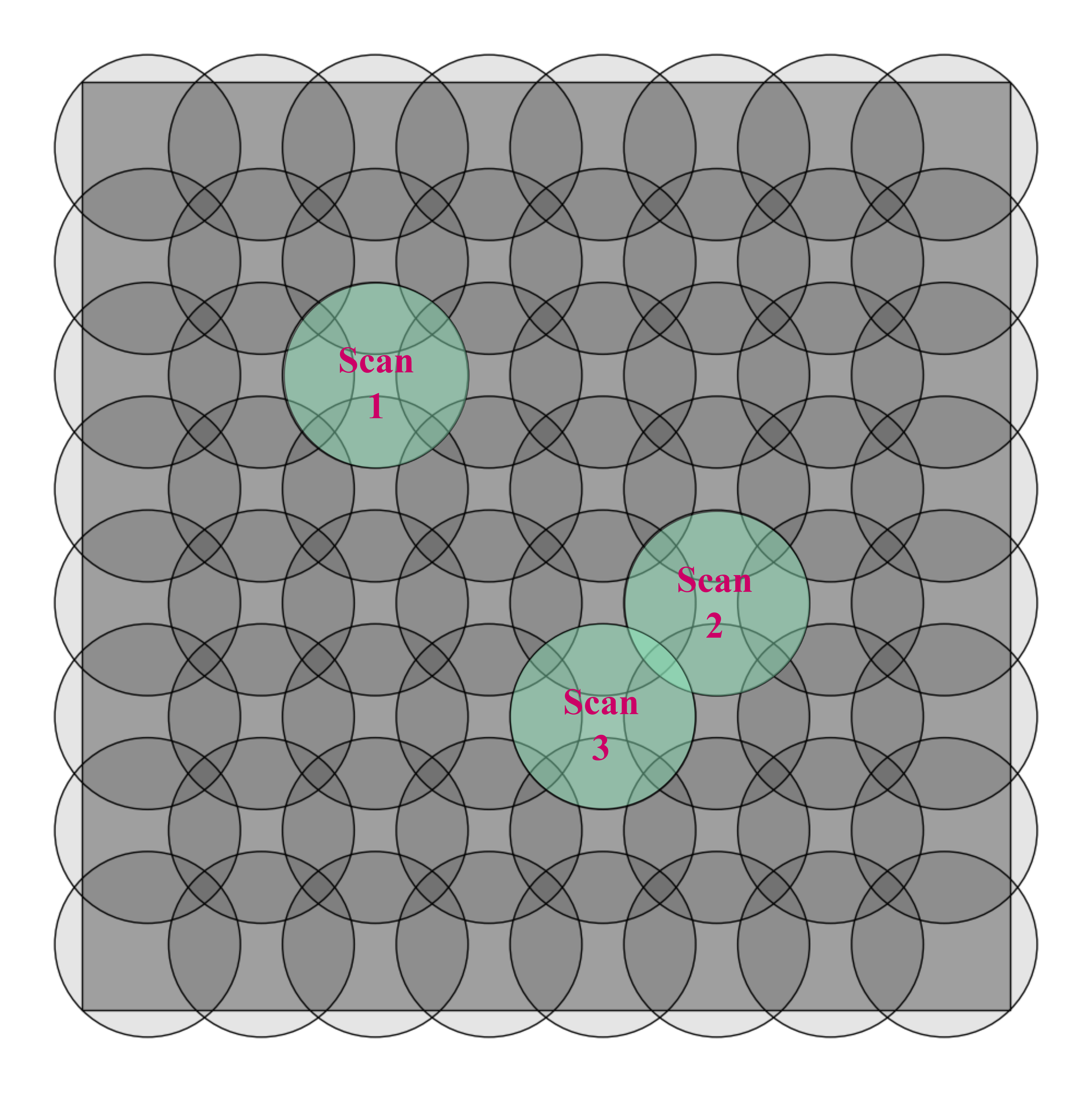}
    \caption{Search region with $N_{a}\times{}N_{a}$ overlapping cells to ensure 100\% coverage of the space. Agent action $\boldsymbol{a}_{k}$ is mapped to a cell with defined azimuth $\psi$ and elevation $\theta$ angles.}
    \label{fig:cell_map}
\end{figure}

A discrete action space similar to \cite{de2023resources} was selected with agent output normalized to generate an azimuth and elevation command associated with a cell in the area of interest. This is shown in Figure \ref{fig:cell_map}.
The number of discrete scan sightlines $N_{a}$ is dependent on the instantaneous FoV of the sensor and the total FoV required to be scanned.
\begin{equation}\label{eq:act_1}
    \boldsymbol{a}_{k} \in \mathbb{Z}^2 \;~|~\; 0 \leq \boldsymbol{a}_{k} < N_a,\quad\textrm{where}\;
    N_a =  \left\lceil\frac{\mathrm{IFoV}}{ \frac{\sqrt{2}}{2}\mathrm{FoV}}\right\rceil 
\end{equation}
The action selected is mapped to an azimuth ($\psi$) and elevation ($\theta$) angle command used to move the Stone Soup sensor sightline.
\begin{equation}\label{eq:actionToBearings}
    [\psi, \theta] = \frac{\pi}{3}
    \left(
    \frac
        {2 \boldsymbol{a}_{k} }
        { N_a - 1}
    -1
    \right) \in [-\frac{\pi}{3}, \frac{\pi}{3}]
\end{equation}
A continuous action space with $\boldsymbol{a}_{k}=c[\psi,\theta]$ would provide the agent more flexibility in selecting dwell centres but would encounter the limitations of the Stone Soup sensor components discussed earlier.


\subsection{Observation Space}
\label{sect:method_observation_space}
The observation space is represented as a continuous set $\boldsymbol{s}_{k}\in\mathcal{O}$ with two components. The first component is a combination of the track list state $\hat{\mathbf{X}}_{k}$ and covariances $\mathbf{P}^{i}_{k}$. The track list information is converted to the observation state vector $\boldsymbol{s}_{TL}\in\mathbb{R}^{N_{track}\times{}7}$ shown below where $N_{track}$ is a parameter that limits the maximum number of tracks in the track list.  If the estimated number of tracks $N_{est}<N_{track}$ then zero-padding is used to retain the input structure. These zeros are then masked upon input into the agent. The track list position information is transformed to a spherical representation since actions are to be outputted in a bearing and elevation form with velocity estimates kept as a Cartesian representation. The final element features the Frobenius norm of the $i^{\textrm{th}}$ track covariance providing the agent with the current uncertainty in the track.
\begin{equation}\label{eq:env_track_list_state}
    \boldsymbol{s}_{TL,i} = \begin{bmatrix}        ||\hat{\boldsymbol{p}_i}|| & \hat{\psi} & \hat{\theta} &\hat{v}_{x_i} & \hat{v}_{y_i} &\hat{v}_{z_i} &||\mathbf{P}_i||_\mathrm{frob}\end{bmatrix}
\end{equation}The second component of the observation state is the scan history represented as a rasterized value map as per \cite{ewers2025multi}.
\begin{equation}
    \boldsymbol{s}_{TL} = ([\boldsymbol{s}_{TL,0},\cdots,\boldsymbol{s}_{TL,N_\mathrm{est}}]~||~\mathbf{0}^{(N_\mathrm{track}-N_\mathrm{est}) \times 7})^T
\end{equation}

Since Stone Soup outputs the track list, a wrapper function is required to form the observation state required as an output from the Gymnasium environment at each time step.

The scan history is the rasterized scan value map defined by 
\begin{equation}
    \boldsymbol{s}_{SH} = \left\{ SSV([\psi, \theta],t) \in \mathbb{R}^{48 \times 48} ~|~ \psi, \theta \in \left[-\frac{\pi}{3}, \operatornamewithlimits{\cdots}_{N-2}, \frac{\pi}{3}\right]\right\}
\end{equation}
such that $\boldsymbol{s}_{SH}$ is an image of dimension $(1,48,48)$.

\subsection{Rewards}
The reward function is defined with two components, one relating to the reduction in uncertainty of the tracks that have associated detections and the second promoting the minimisation of the scan value. The latter results from improved search performance with move of the area of interest scanned. The total reward is constructed as follows.
\begin{equation}
\begin{split}
    r_k = \sum \left\{||\mathbf{P}_k||_\mathrm{fro} - ||\mathbf{P}_{k-1}||_\mathrm{fro} ~|~\mathcal{D}_k \bigcup \mathcal{D}_{k-1}\right\}\\ - SSV([\psi, \theta], k)
\end{split}
\end{equation}
Here $\mathcal{D}_k$ is the set of detections at the current timestep $k$ and $\mathbf{P}_k$ is the error covariance matrix of the associated track for the set of detections $\mathcal{D}_k$. The Stone Soup \textit{SumOfCovarianceNorms} metric is used to compute the Frobenius norm used in the reward function and as a metric for analysing agent performance.
\subsection{Network Design}
\label{sect:network_design}
The sensor management problem utilising an MTT algorithm presents some non-trivial challenges for constructing the agent architecture. The track list $\hat{\mathbf{X}}_{k}$ generated by the MTT algorithm will be a tuple, or potentially random finite set (RFS), with varying dimension based on the scenario timestep, sensor parameters and performance of the MTT algorithm. Depending on the neural network layers desired a number of solutions exist. The first is to pad the track list with zeros up to a predefined maximum dimension, and then to perform masking so that the agent knows to ignore zeroed elements. This approach is useful for implementation of MLP and CNN layers which require a fixed size input. This is the approach taken in the sample implementation with the track list component of the environment state $\boldsymbol{s}_{TL}$ featuring the $n$ current track states as defined by (\ref{eq:env_track_list_state}). The track list state captures the current best estimate of target states at the current timestep. As such, this data is non-temporal. Bi-directional recurrent neural network layers such as BiLSTM and BiGRU \cite{Cahuantzi2023} have been shown to perform well with non-temporal input states \cite{Wang2020}. Gated recurrent unit (GRU) implementations are a simpler alternative to full LSTM or recurrent layers and would be more suitable for lower dimension data. The alternative approach is to make use of transformer architectures with the self-attention mechanism \cite{vaswani2017attention}. Self-attention layers enable the agent to prioritise the importance of parts of the input state to other elements. For the track list state, this would enable the agent to capture any relationships between tracks captured in the observation state. Multi-head self-attention (MHSA) expands this process to enable multiple relationships to be learned between elements of the input state. Three network architectures using a basic \textit{flattening} layer, BiGRU layer and BiGRU with prior MHSA are implemented in the sample problem. The BiGRU+MHSA network requires the recurrent layer to transform the sequence output from the MHSA block to a vector for latent space concatenation with information generated from the CNN. The scan value map $\boldsymbol{s}_{SH}$ representing the search information the Radar has generated as an image. To perform the feature extraction from the scan value map, NatureCNN, a convolutional neural network (CNN) architecture proposed in \cite{mnih2015human} and shown to effectively extract information from 2D game states is utilised on the updated image. The two streams of feature extraction from the track list $\boldsymbol{z}_{TL}\in\mathbb{R}^{64}$ and scan value map components $\boldsymbol{z}_{SH}\in\mathbb{R}^{128}$ are concatenated in the latent space $\boldsymbol{z}$ before being fed into the core policy comprised of a MLP.
\begin{figure*}[tb!]
    \centering
    \subfloat[
                $\boldsymbol{s}_{TL}$ is flattened and concatenated with the NatureCNN output before being inputted into the core network.
                \label{fig:rl_architecture_flat}
            ]{
                \includegraphics[width=.25\linewidth,origin=c]{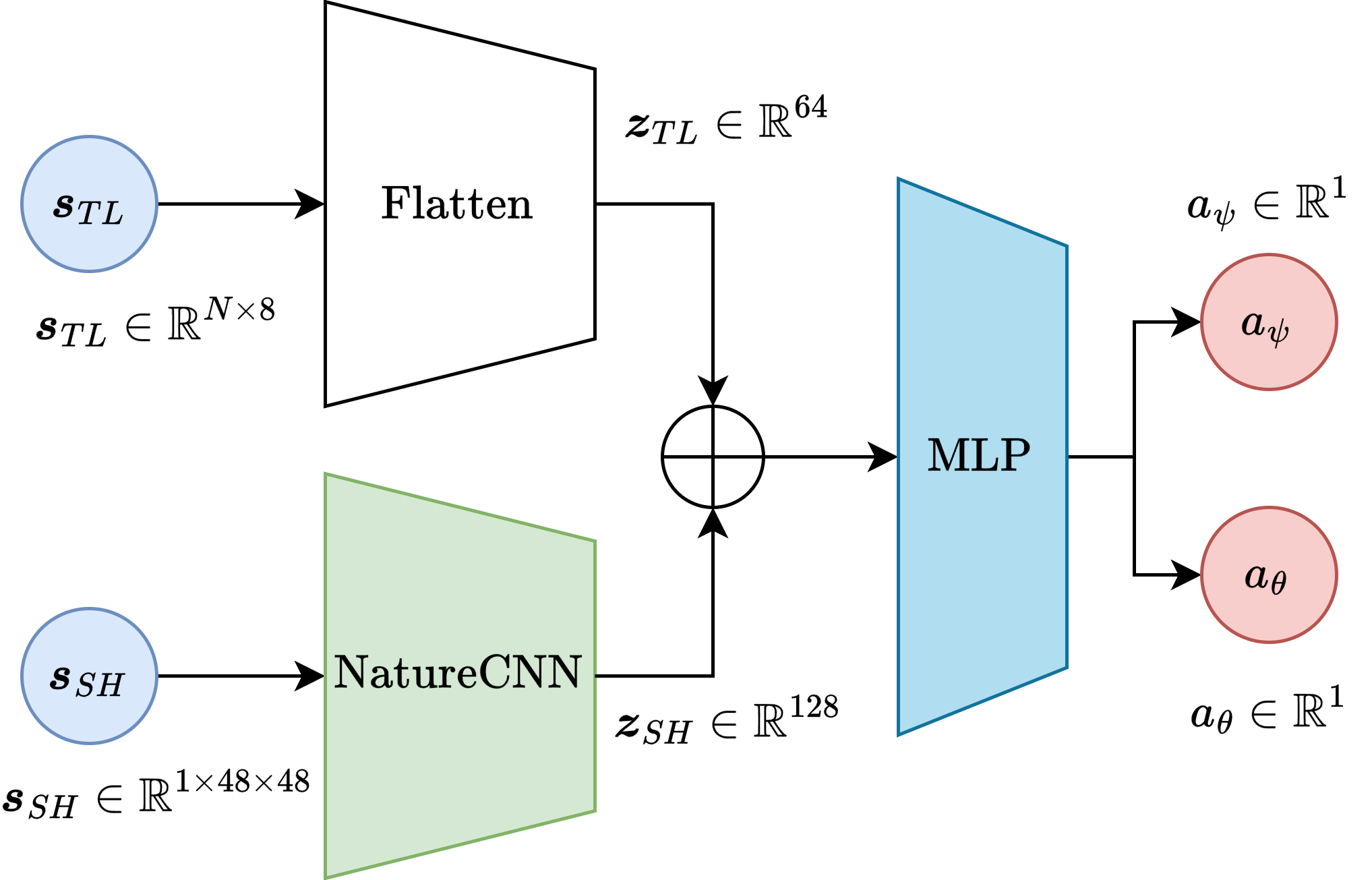}
            }
    \hfill
    \subfloat[
                BiGRU architecture where $\boldsymbol{s}_{TL}$ is passed into the recurrent network sequentially outputting a fixed-length latent representation.
                \label{fig:rl_architecture_rnn}
            ]{
                \includegraphics[width=.25\linewidth,origin=c]{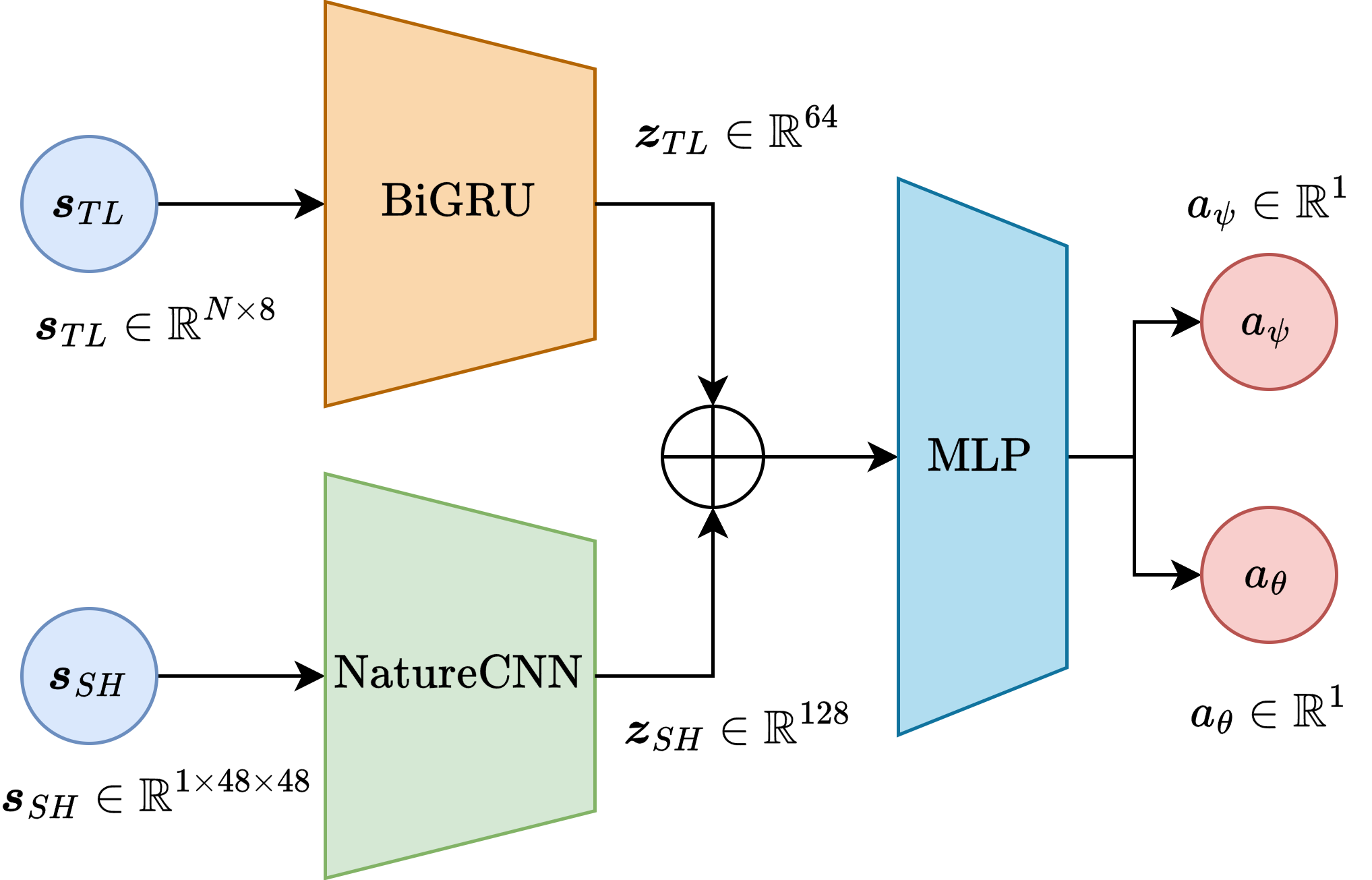}
            }
    \hfill
    \subfloat[
                Multi-headed self attention is used to highlight important tracks from $\boldsymbol{s}_{TL}$. The BiGRU module is $50\%$ of the size as the standalone variant. The recurrent layer is required to transform the sequence information from the MHSA layers into a vector representation for concatenantion with the NatureCNN latent representation.
                \label{fig:rl_architecture_rnn_mhsa}
            ]{
                \includegraphics[width=.37\linewidth,origin=c]{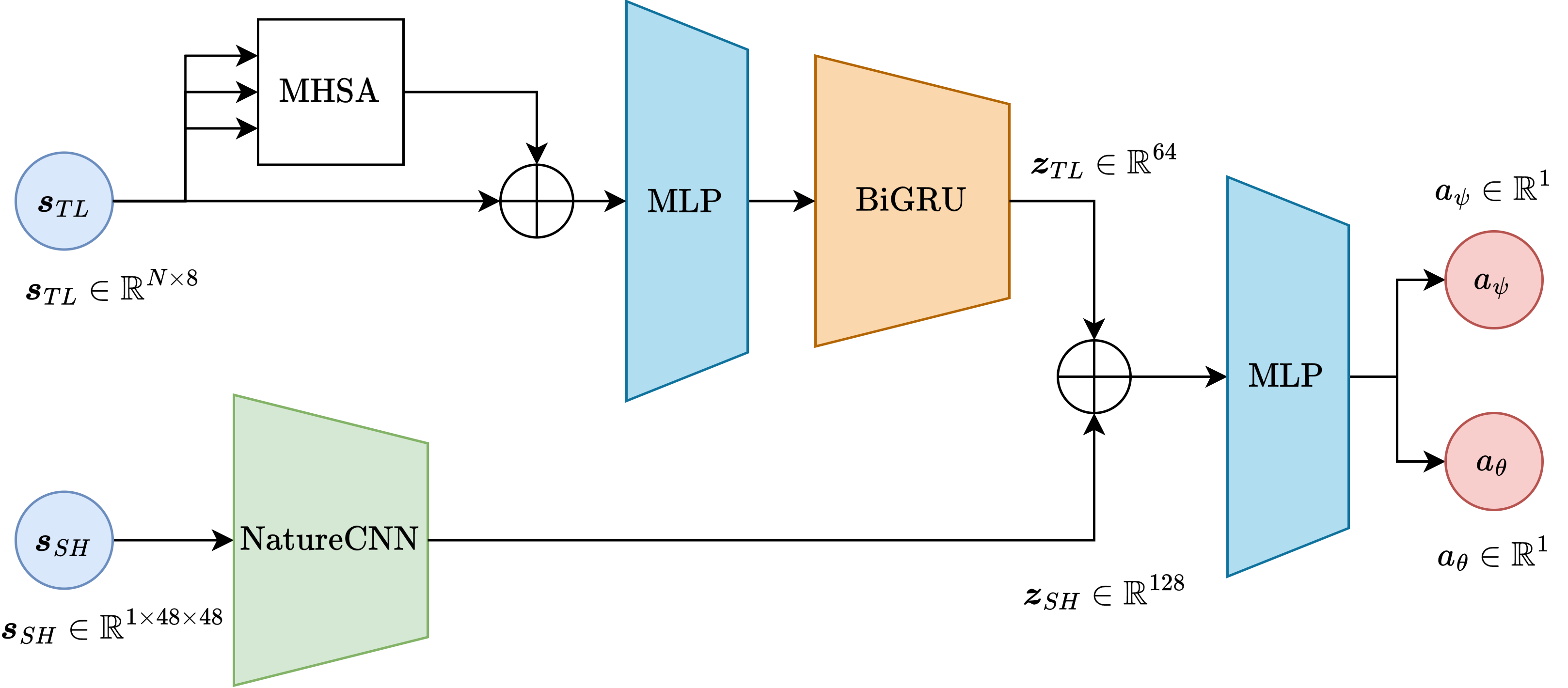}
            }
    \caption{Policy architectures for sample sensor management problem. A NatureCNN \cite{mnih2015human} extracts features from the scan-history. A flattening, recurrent or multi-headed self attention network is implemented to transform the track-list information for concatenation into the MLP component of the policy.}
    \label{fig:enter-label}
\end{figure*}

\section{Results}
\label{sect:results}

\begin{figure*}[tb!]
    \centering    
    \subfloat[Mean $x$-position covariance for tracks within each policy simulation.\label{fig:results_x_cov}]{\includegraphics[width=0.24\linewidth]{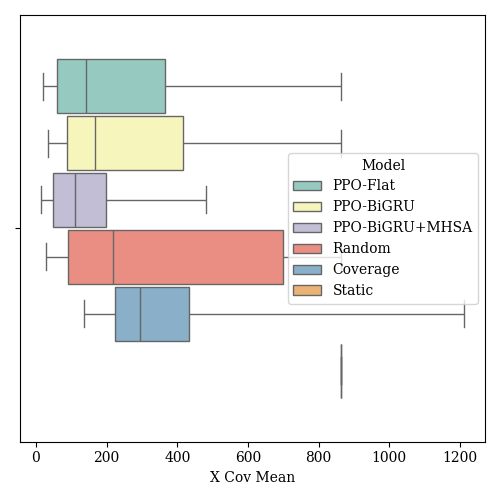}}
    \hfill
    \subfloat[Mean $y$-position covariance for tracks within each policy simulation.\label{fig:results_y_cov}]{\includegraphics[width=0.24\linewidth]{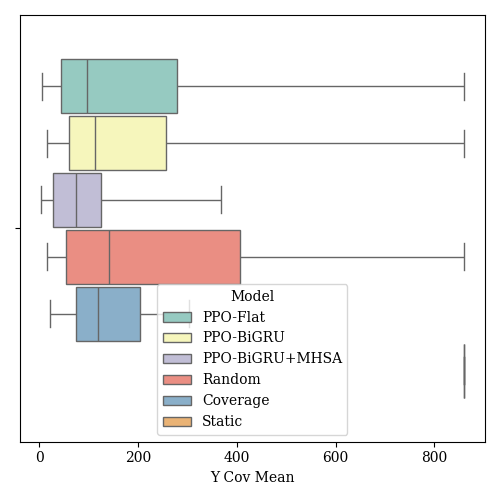}}
    \hfill
    \subfloat[Mean $z$-position covariance for tracks within each policy simulation.\label{fig:results_z_cov}]{\includegraphics[width=0.24\linewidth]{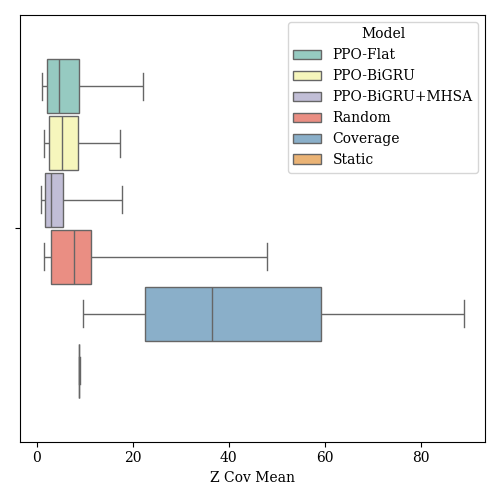}}
    \hfill
    \subfloat[Mean covariance norm for track position and velocity within each policy simulation.\label{fig:results_cov_norm}]{\includegraphics[width=0.24\linewidth]{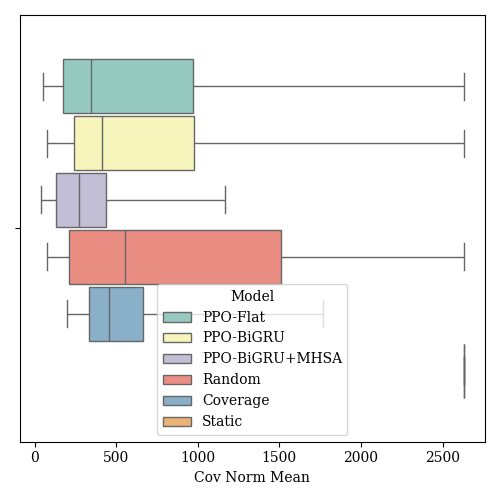}}
    \caption{Mean position covariance and full-state mean covariance norm for each policy implemented in the search and track sensor management simulation. The PPO-BiGRU with multi-head self-attention produces the lowest covariance estimates out of all policies implemented.}
    \label{fig:results_cov_norm_mean_all}
\end{figure*}
\begin{figure*}[tb!]
    \centering    
    \subfloat[GOSPA distance metric for each policy.\label{fig:results_gospa_distance}]{\includegraphics[width=0.24\linewidth]{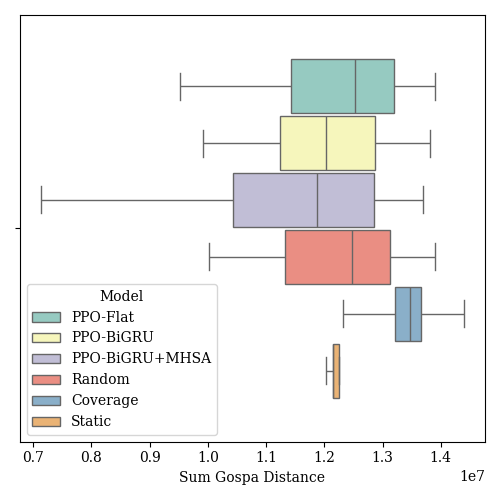}}
    \hfill
    \subfloat[GOSPA localisation metric for each policy.\label{fig:results_gospa_localisation}]{\includegraphics[width=0.24\linewidth]{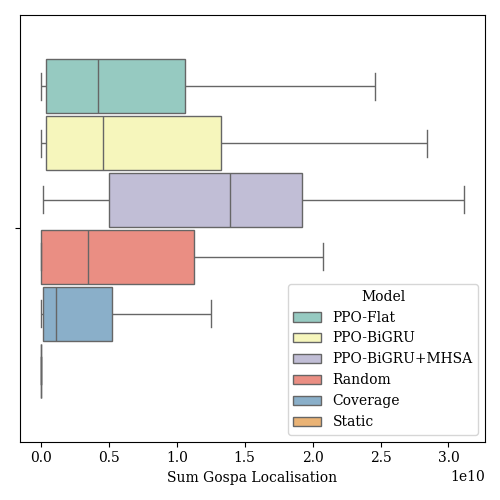}}
    \hfill
    \subfloat[GOSPA false alarm metric for each policy.\label{fig:results_gospa_false}]{\includegraphics[width=0.24\linewidth]{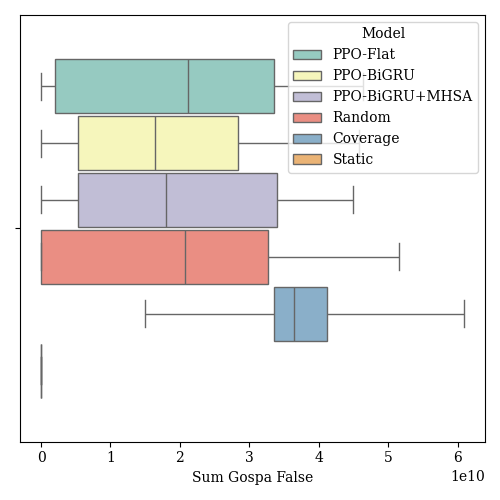}}
    \hfill
    \subfloat[GOSPA missed tracks metric for each policy.\label{fig:results_gospa_missed}]{\includegraphics[width=0.24\linewidth]{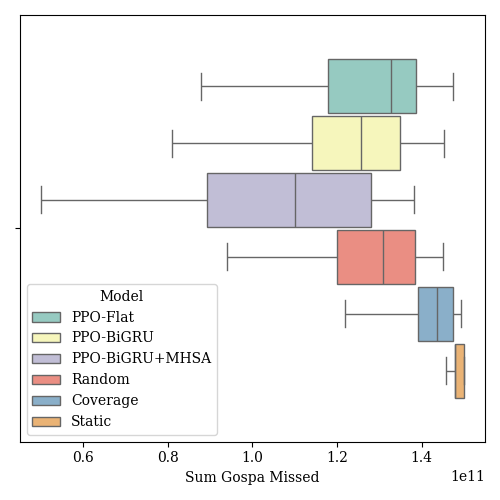}}
    \caption{GOSPA distance and consituent GOSPA metrics \autoref{fig:results_gospa_distance} \cite{Rahmathullah2017} for each policy implemented. GOSPA is calculated using the \textit{GOSPAMetric} metric generator component in Stone Soup.}
    \label{fig:results_gosap_all}
\end{figure*}

\subsection{Example implementation}
\label{sect:results_experimental_setup}
The three agent architectures defined above are compared with three simple if not representative search policies. These policies are as follows.
\begin{itemize}
    \item \textit{Random} - the action is uniformly sampled at each time step from the possible action space
    \item \textit{Static} - a single initial action is uniformly sampled from the action space and kept constant
    \item \textit{Coverage} - each line is scanned sequentially in a common search technique
\end{itemize}

\begin{table}[htbp]
\caption{Agent Network Parameters for the Flatten, BiGRU and multi-head Self-Attention BiGRU architectures.}
\begin{center}
\begin{tabular}{|l|l|c|}
\hline
\textbf{Policy}&\textbf{Component}&\textbf{Dimensions}\\
\hline
PPO-Flatten&NatureCNN Output&$1\times128$\\
\hline
PPO-Flatten&MLP&$2\times128$\\
\hline
PPO-BiGRU&NatureCNN Output&$1\times128$\\
\hline
PPO-BiGRU&BiGRU Hidden Size&$64$\\
\hline
PPO-BiGRU&MLP&$2\times128$\\
\hline
PPO-BiGRU-MHSA&NatureCNN Output&$1\times128$\\
\hline
PPO-BiGRU-MHSA&BiGRU Hidden Size&$32$\\
\hline
PPO-BiGRU-MHSA&MLP&$2\times128$\\
\hline
\end{tabular}
\label{tab:network_params}
\end{center}
\end{table}

\subsection{Summary of example results}
\label{sect:results_track}
The agent performance was analysed over 100 simulations with track-to-truth, GOSPA and mean covariance norms saved from the Stone Soup metric components to compare performance. The mean track-to-truth values shown in Table \ref{tab:meangospa} show good performance of all agent architectures although the random search algorithm performs best since it does not prioritise any track revisits or scanning of unsearched areas. This can be seen in the mean covariance norm results in Table \ref{tab:meangospa}, showing poor confidence in the track estimates with the BiGRU with multi-head self-attention architecture performing best. Full metrics are provided in Figures \ref{fig:results_cov_norm_mean_all} and \ref{fig:results_gosap_all}.

\begin{table}[htbp]
\caption{Track Metrics for each agent and baseline search policy.}
\begin{center}
\begin{tabular}{|l|c|c|}
\hline
\textbf{Policy}&\textbf{Mean T2T}&$||\bar{\mathbf{P}}||_{\textrm{frob}}\;[\textrm{m}]$\\
\hline
Coverage&$0.942$&$574.27\pm{}424.53$\\
\hline
Random&$1.585^{\ast}$&$948.15\pm{}921.87$\\
\hline
Static&$0.288$&$2630.72\pm{}2358.40$\\
\hline
PPO-Flatten&$1.407$&$790.63\pm{}852.00$\\
\hline
PPO-BiGRU&$1.430$&$730.56.27\pm{}744.81$\\
\hline
PPO-BiGRU-MHSA&$1.152$&$461.98\pm{}573.50^{\ast}$\\
\hline
\multicolumn{3}{l}{$^{\mathrm{\ast}}$Indicates best performance in each metric.}
\end{tabular}
\label{tab:meangospa}
\end{center}
\end{table}
The example implementation for a search and track sensor management problem has demonstrated the utility of the Stone Soup framework for use as a feature extractor within an RL agent. Results from this simulation show that the multi-head self-attention operation improves the agent's performance with the lowest GOSPA distance overall as shown in \autoref{fig:results_gosap_all} and lowest Frobenius norm of the covariance in Table \ref{tab:meangospa}.

\section{Conclusions and Future Work}
\label{sect:conclusion_future_work}

A sample architecture for integrating Stone Soup tracker components within a Gymnasium environment for use in reinforcement learning scenarios has been presented and applied to a sensor management problem. The approach has demonstrated the efficient use of Stone Soup tracking components within the Stable Baselines3 environment to utilise the wide array of policy methods and neural network layers available from PyTorch. The general approach presented can be adapted for a variety of autonomy or sensor management problems which require a capable feature extractor to generate information for a RL agent although some level of customisation is still required, dependent on the type of problem being investigated. Current limitations in the Stone Soup library constrain the integration within Gymnasium, and different studies may require additional components and wrapper functions to enable conversion between Stone Soup objects and the required structures for Stable Baselines3. For the Radar management problem specifically, a few limitations should be highlighted. At present and although supported, Gymnasium does not provide full support for multi-agent reinforcement learning environments, so alternative packages such as PettingZoo \cite{terry2021pettingzoo} would need to be utilised in scenarios with multiple independent sensors. The current track list state $\boldsymbol{s}_{TL}$ holds the present target information, but an agent with recurrent, convolutional or self-attention layers should be capable of projecting ahead using past track lists to extract temporal information about the targets \cite{milan2017online}. Upgrading the Stone Soup MTT to utilise tracking algorithms such as the state-of-the-art trajectory-PMBM \cite{garcia2020trajectory}, would provide a \textit{trajectory list} with further scope for feature extraction. Stone Soup is yet to fully incorporate PMBM approaches but incorporation within the presented architecture should not require significant changes if it could be implemented within Stone Soup.

\section*{Acknowledgment}
Thank you to David Cormack for his insight into the project. Additional thanks to Paul Thomas for organising the special session on \textit{Applications of Stone Soup} at FUSION 2025.


\bibliography{references.bib}
\bibliographystyle{ieeetr}

\end{document}